\pdfoutput=1

\documentclass[10pt,twocolumn,letterpaper]{article}

\usepackage[pagenumbers]{wacv}

\newenvironment{acks}{\par\noindent\textbf{Acknowledgments.}}{\par}

\usepackage{tikz, siunitx, contour}
\usepackage{booktabs, multirow}
\newcommand{\outlinedcolor}[2]{\contour{black}{\textcolor{#1}{#2}}}

\definecolor{wacvblue}{rgb}{0.21,0.49,0.74}
\usepackage[pagebackref,breaklinks,colorlinks,allcolors=wacvblue]{hyperref}

\def\wacvPaperID{11} 
\def\confName{WACV}
\def\confYear{2027}

\title{GOOSELoc: A Multimodal Benchmark for Ground-to-Aerial Localization in Unstructured Off-Road Terrain}

\author{
Fabian Deuser$^{1,3}$ \quad Peter Mortimer$^{2}$ \quad Robin-Nico Kampa$^{1}$ \quad Anna Bößendörfer$^{1}$ \\
Mirko Mählisch$^{2}$ \quad Martin Werner$^{3,4}$ \quad Norbert Oswald$^{1}$ \\[0.5em]
\parbox{0.95\textwidth}{\centering\small
$^{1}$Institute of Distributed Intelligent Systems (ETTI~2), University of the Bundeswehr Munich, Neubiberg, Germany \\
$^{2}$Institute for Autonomous Driving (LRT~8.1), University of the Bundeswehr Munich, Neubiberg, Germany \\
$^{3}$Technical University of Munich, Germany; TUM School of Engineering and Design, Department of Aerospace and Geodesy \\
$^{4}$TUM School of Computation, Information and Technology; Munich Data Science Institute (MDSI) \\[0.3em]
{\tt \{fabian.deuser, peter.mortimer, robin-nico.kampa, anna.boessendoerfer, mirko.maehlisch, norbert.oswald\}@unibw.de, martin.werner@tum.de}
}
}
\begin{document}
\maketitle

\begin{figure*}[htbp]
    \definecolor{surround_front}{RGB}{178, 178, 255}
    \definecolor{surround_right}{RGB}{178, 216, 216}
    \definecolor{surround_back}{RGB}{216, 178, 216}
    \definecolor{surround_left}{RGB}{255, 216, 178}
    \definecolor{lidar}{RGB}{255, 80, 80}
    \definecolor{aerial}{RGB}{23, 156, 125}
    \definecolor{star}{RGB}{240, 199, 0}
    \centering
    \includegraphics[width=.4\textwidth]{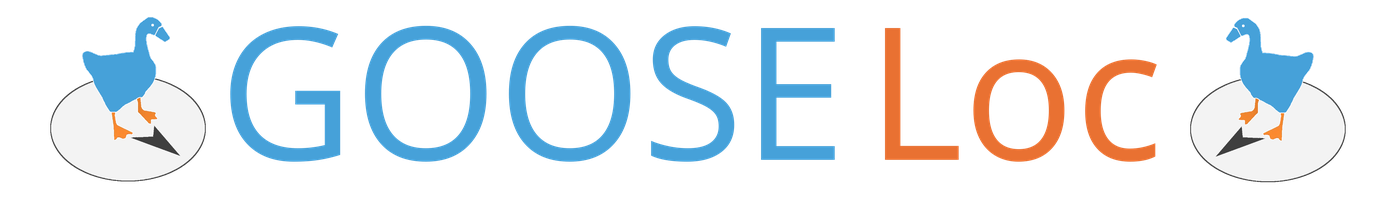}
    \begin{tikzpicture}
    \node[anchor=south west,inner sep=0] (img) at (0,0)
    {\includegraphics[width=1.0\textwidth, trim={0mm 0mm 0mm 0mm},clip]{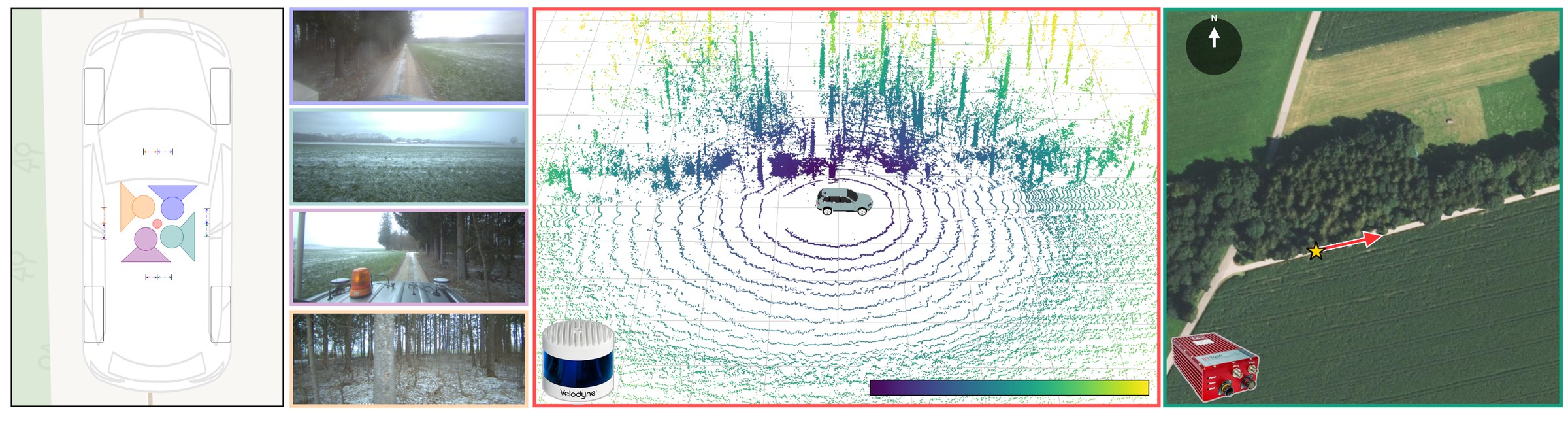}};
    
    \begin{scope}[x={(img.south east)}, y={(img.north west)}]
    
        \node[anchor=west,  draw=surround_front, rounded corners=1pt, fill=lightgray,fill opacity=0.75,text opacity=1,inner sep=2pt] at (0.19,0.935) {\tiny front};
        \node[anchor=west,  draw=surround_right, rounded corners=1pt, fill=lightgray,fill opacity=0.75,text opacity=1,inner sep=2pt] at (0.19,0.7) {\tiny right};
        \node[anchor=west, draw=surround_back, rounded corners=1pt, fill=lightgray,fill opacity=0.75,text opacity=1,inner sep=2pt] at (0.19,0.46) {\tiny back};
        \node[anchor=west, draw=surround_left, rounded corners=1pt, fill=lightgray,fill opacity=0.75,text opacity=1,inner sep=2pt] at (0.19,0.22) {\tiny left};
    
        \node[anchor=west,text opacity=1] at (0.055,0.675) {\tiny \SI{0.18}{\meter}};
        \node[anchor=west,text opacity=1] at (0.095,0.675) {\tiny \SI{0.20}{\meter}};
        \node[anchor=west,text opacity=1] at (0.02,0.5) {\tiny \SI{0.22}{\meter}};
        \node[anchor=west,text opacity=1] at (0.02,0.46) {\tiny \SI{0.20}{\meter}};
        \node[anchor=west,text opacity=1] at (0.055,0.315) {\tiny \SI{0.15}{\meter}};
        \node[anchor=west,text opacity=1] at (0.095,0.315) {\tiny \SI{0.19}{\meter}};
        \node[anchor=west,text opacity=1] at (0.13,0.5) {\tiny \SI{0.20}{\meter}};
        \node[anchor=west,text opacity=1] at (0.13,0.46) {\tiny \SI{0.17}{\meter}};

        \node[anchor=west,text opacity=1] at (0.55,0.125) {\tiny \SI{0}{\meter}};
        \node[anchor=west,text opacity=1] at (0.7,0.125) {\tiny \SI{35}{\meter}};
    
    \end{scope}
    \end{tikzpicture}
    \caption[]{
    \textbf{GOOSELoc Dataset Sample.} The 4 roof-mounted cameras \outlinedcolor{surround_front}{$\blacksquare$} \outlinedcolor{surround_right}{$\blacksquare$} \outlinedcolor{surround_back}{$\blacksquare$} \outlinedcolor{surround_left}{$\blacksquare$} each have a $2448\times1000$\,px resolution and a \SI{94.71}{\degree} horizontal field of view after rectification, giving us a \SI{360}{\degree} coverage of the surroundings. 
    This can be fused with a sweep of the roof-mounted 128-beam rotating LiDAR scanner \outlinedcolor{lidar}{$\blacksquare$}. 
    The Bavarian orthophotos \outlinedcolor{aerial}{$\blacksquare$} are taken on a clear day in the summer and have a temporal mismatch to the ground-view images recorded during winter.
    The ground truth global location and heading \outlinedcolor{star}{$\bigstar$} are obtained by an Oxford RT3000v3 INS system with differential RTK-GNSS corrections using the NTRIP protocol to receive accurate readings in forested areas.
    }
    \label{fig:gooselocsample}
\end{figure*}

\begin{abstract}
Cross-view pose estimation estimates a ground camera's geographic pose by matching onboard sensor data with geo-referenced overhead imagery. This is particularly relevant when GNSS is unreliable, for example under forest canopy or in urban canyons. Existing metric cross-view benchmarks, such as VIGOR and cross-view KITTI, are dominated by urban and road scenes with strong human-made structure without off-road terrain. We introduce GOOSELoc, a multimodal benchmark for ground-to-aerial localization in unstructured environments, including forests, fields, and rural tracks. GOOSELoc comprises 9,356 scenes recorded across all four seasons around Munich, Germany, pairing geo-referenced aerial orthophotos with surround-view RGB imagery and 3D LiDAR point clouds. Paved/unpaved labels enable analyzing the impact of terrain structure, while the seasonal recordings allow studying the appearance gap to mostly summer-captured orthophotos. To assess the transferability of existing methods, we adapt and fine-tune four state-of-the-art cross-view localization baselines on GOOSELoc. Despite task-specific fine-tuning, the performance degrades substantially in off-road environments. We therefore propose a multimodal baseline combining surround-view imagery and LiDAR geometry, improving localization performance and providing a stronger reference for future work. Data and baselines are released at \url{https://gooseloc.github.io}.
\end{abstract}    
\section{Introduction}
\label{sec:intro}
Accurate geo-localization is one of the cornerstones of modern navigation and underpins the safe transportation of goods and people. Systems deployed in cars, trucks, and UAVs rely predominantly on Global Navigation Satellite Systems (GNSS). However, GNSS signals degrade systematically in environments where direct sky visibility is obstructed: urban canyon effects limit accuracy in dense cities~\cite{brosh2019accurate}, and tree canopies together with uneven terrain cause comparable or worse degradation in forested and off-road areas~\cite{wing2005consumer,zhu2018gnss}. Robust backup systems are therefore required, particularly for the off-road domain, where these conditions are the norm rather than the exception. 

Cross-view geo-localization~\cite{lin2013cross} is a computer-vision-based alternative: it estimates a ground sensor's pose by relating its observations to globally available, geo-referenced aerial imagery, yielding absolute positions. Existing benchmarks address two task formulations. Retrieval-oriented datasets such as CVUSA~\cite{workmann2015cvusa,zhai2017predicting} and CVACT~\cite{liu2019lending} pair ground panoramas with aerial tiles and treat localization as nearest-neighbor matching against a tile database, where the recoverable position is bounded by the tile size. Metric pose estimation is targeted by VIGOR~\cite{zhu2021vigor}, which additionally supports retrieval, and by the cross-view extension of KITTI~\cite{geiger2013vision} introduced by Shi et al.~\cite{shi2022accurate}, both targeting sub-tile localization at the meter level.
These benchmarks share a limitation that reflects their origin: all target urban, road-centric settings, with imagery captured from street-view panoramas or road-going vehicles along paved roads. Even datasets that include rural imagery, CVUSA for instance, capture it along highways and paved roads. Off-road perception itself is well studied~\cite{wigness2019rugd,jiang2021rellis,mortimer2024goose}, but almost exclusively for segmentation and navigation rather than ground-to-aerial localization. This setting also imposes a fundamentally different appearance regime. Whereas urban and road scenes provide stable, human-made cross-view anchors such as buildings, lane markings and signage~\cite{deuser2023sample4geo}, unstructured terrain is dominated by vegetation whose appearance varies strongly with the season and shifts continuously between recordings, a well-documented source of degradation for place recognition and pose estimation~\cite{maddern20171,sattler2018benchmarking}. Orthophotos aggravate this mismatch: flown once or twice a year under clear-weather summer conditions, they freeze a single vegetation state, and even moderate ground-aerial temporal offsets are known to degrade cross-view performance~\cite{deuser2025temporal}.

We address this gap with GOOSELoc, a multimodal benchmark for ground-to-aerial localization in unstructured off-road terrain. GOOSELoc builds on the German Outdoor and Offroad Dataset (GOOSE)~\cite{mortimer2024goose}, originally introduced for outdoor and off-road semantic segmentation, and uses accurately localized multimodal recordings captured across all four seasons around Munich, Germany. We filter the recordings for positional accuracy and make the following contributions:
\begin{itemize}
\item We introduce GOOSELoc, a multimodal benchmark for metric ground-to-aerial localization in unstructured off-road terrain, comprising 9,356 scenes paired with geo-referenced 20\,cm ground sample distance (GSD) aerial orthophotos, see \Cref{fig:gooselocsample}.
\item We provide synchronized surround-view RGB imagery and 3D LiDAR point clouds for each scene, together with paved/unpaved labels and four-season coverage, enabling analyses of terrain structure and seasonal appearance gaps.
\item We adapt and fine-tune CCVPE~\cite{xia2023convolutional}, the work of Shi et al.~\cite{shi2023boosting}, HC-Net~\cite{wang2023fine} and FG\textsuperscript{2}~\cite{xia2025fg}, as state-of-the-art cross-view localization methods, and show that they degrade substantially on GOOSELoc. We further propose an RGB-LiDAR fusion baseline that improves over the fine-tuned single-modality baseline.
\end{itemize}
\section{Related Work}
\label{sec:relwork}
\paragraph{Cross-View Datasets}
Cross-view geo-localization was introduced by Lin et al.~\cite{lin2013cross} as a feature-translation problem between ground-level and overhead imagery. The field then consolidated around a set of retrieval benchmarks: CVUSA~\cite{workmann2015cvusa,zhai2017predicting}, the cross-area splits of Vo et al.~\cite{vo2016localizing}, and CVACT~\cite{liu2019lending} all match (near-)centered street-view panoramas against overhead tiles, while University-1652~\cite{zheng2020university} replaces the ground side with UAV queries. VIGOR~\cite{zhu2021vigor} adds a fine-grained benchmark requiring meter-level regression within the retrieved patch, and Shi et al.~\cite{shi2022accurate} extend KITTI~\cite{geiger2013vision} to a harder 3-DoF setting in which only a forward-facing camera is available. All of these remain urban or automotive.
\paragraph{Cross-View Localization Methods}
Methods differ mainly in how precisely they localize. The retrieval line learns one global descriptor per view, so a ground query and its true overhead tile become nearest neighbors in a shared embedding. Early work paired this idea with task-specific modules for the panorama--aerial geometry gap: NetVLAD aggregation~\cite{hu2018cvmnet,arandjelovic2016netvlad}, a polar transform with re-weighting of informative regions~\cite{shi2019safa}, and a ViT~\cite{dosovitskiy2020image} attending twice to zoom into relevant patches~\cite{zhu2022transgeo}. Sample4Geo~\cite{deuser2023sample4geo} drops these modules: a ConvNeXt encoder~\cite{liu2022convnet} trained with a symmetric InfoNCE loss~\cite{oord2018representation} and hard negatives from geography and embedding space set the retrieval state of the art at the time and transferred better to unseen regions. Others add a bird's-eye view (BEV) of the ground image as a second, geometrically aligned cue, either warping the panorama directly~\cite{ye2024cross} or reusing a pyramid-occupancy transform pre-trained for semantic BEV prediction~\cite{shore2024bev}.
\begin{table*}[ht!]
\centering
\small
\setlength{\tabcolsep}{6pt}
\resizebox{\textwidth}{!}{%
\begin{tabular}{lcccccc}
\toprule
Dataset & Size & Ground view & Modalities & Environment & Seasonal gap & Localization \\
\midrule
CVUSA~\cite{workmann2015cvusa,zhai2017predicting} & 44.4k & Panorama & RGB & Urban / rural & No & Retrieval \\
CVACT~\cite{liu2019lending} & 137.2k & Panorama & RGB & Urban & No & Retrieval \\
VIGOR~\cite{zhu2021vigor} & 105.2k & Panorama & RGB & Urban & No & Retrieval + Metric \\
Cross-View KITTI~\cite{shi2022accurate} & 30.2k & Forward & RGB & Urban / rural & No & Metric \\
University-1652~\cite{zheng2020university} & 146.0k & Social Media & RGB & Urban (buildings-only) & No & Retrieval \\
MoViX / TartanDrive\,2.0~\cite{deng2025spatiotemporal,sivaprakasam2024tartandrive} & 4.6k & Forward (video) & RGB & Off-road & Yes & Metric \\
McPed~\cite{yuan2025multi} & 29.9k & Forward & RGB\,+\,LiDAR & Field / urban & No & Metric \\
\midrule
\textbf{GOOSELoc (ours)} & \textbf{9.4k} & \textbf{Surround 360$^\circ$} & \textbf{RGB\,+\,LiDAR} & \textbf{Off-road / rural / urban} & \textbf{Yes} & \textbf{Metric} \\
\bottomrule
\end{tabular}%
}
\caption{GOOSELoc in the context of existing cross-view benchmarks. A GOOSELoc scene comprises four surround images, a LiDAR sweep, and a geo-referenced orthophoto crop. GOOSELoc is the only benchmark pairing surround-view imagery and LiDAR against official orthophotos, and the only one combining paved-to-off-road coverage with a controlled seasonal ground-aerial gap. McPed~\cite{yuan2025multi} is the closest multimodal relative: RGB\,+\,LiDAR from a single forward-facing $79^\circ$ camera on a tracked robot, recorded in clear weather without a seasonal protocol, with SLAM-derived ground truth and per-sequence rather than geographic splits. MoViX shares the off-road setting and a ground-aerial temporal gap but relies on a single forward-facing RGB camera.}
\label{tab:datasets}
\end{table*}
Retrieval returns a tile, not a metric pose. A second line therefore estimates the camera pose inside a known aerial patch. SliceMatch~\cite{lentsch2023slicematch} scores a ground descriptor against pose-dependent aerial descriptors aggregated over horizontal viewing-direction slices, keeping the best 3-DoF candidate. CCVPE~\cite{xia2023convolutional} predicts a dense, orientation-aware location heatmap coarse-to-fine with a translation-equivariant encoder, and HC-Net~\cite{wang2023fine} casts localization as homography estimation through a differentiable spherical transform. FG\textsuperscript{2}~\cite{xia2025fg} matches fine-grained features from a frozen DINOv2 backbone~\cite{oquab2023dinov2}: following camera-to-BEV perception~\cite{philion2020lift,li2024bevformer}, it lifts ground features into a 3D point cloud, learns where along the height to pool them onto a BEV plane, and aligns the result to aerial points via Procrustes.

\paragraph{LiDAR-based and Cross-Modal Localization}
A parallel line localizes from 3D geometry rather than imagery. Building on learned global descriptors for point-cloud place recognition~\cite{uy2018pointnetvlad}, later work crosses modalities: VXP~\cite{li2024vxp} retrieves a camera query against a LiDAR map in a shared embedding, CrossLoc3D~\cite{guan2023crossloc3d} matches ground against aerial point clouds, and L2RSI~\cite{shi2026l2rsi} localizes LiDAR submaps against overhead imagery. Geometry is attractive here because it is sensed actively and therefore independent of illumination, yet all of these operate in urban environments and pair one query modality with one reference, rather than combining LiDAR with imagery on the ground side.

Only recently has cross-view localization moved off-road. Yuan et al.~\cite{yuan2025multi} contribute a dataset covering unstructured terrain, and MoViX~\cite{deng2025spatiotemporal} performs cross-view video localization on TartanDrive 2.0~\cite{sivaprakasam2024tartandrive}. Both, however, observe the scene through a single forward-facing camera, so the surrounding context that carries much of the signal in cross-view matching is never available, and neither combines this view with LiDAR. GOOSELoc targets exactly this gap.
\providecommand{\dataset}{GOOSELoc\xspace}
\section{The GOOSELoc Dataset}
\label{sec:dataset}
\dataset turns the ground-level imagery of the GOOSE
dataset~\cite{mortimer2024goose} into a benchmark for cross-view visual localization in unstructured terrain, as can be seen in \Cref{fig:gooselocsample}. Each of the 9{,}356 ground scenes is paired with a co-registered aerial orthophoto and an absolute ground-truth pose, so that the task is to localize a vehicle from a single ground-level capture against a top-down aerial map. Unlike existing geo-localization benchmarks, which are dominated by structured urban street scenes, \dataset covers a near-even mix of paved roads and unstructured terrain---forest tracks, field paths, and rural roads---and spans a full calendar year, exposing methods to strong seasonal and appearance change. We compare it in \Cref{tab:datasets} with other currently available datasets.
\subsection{Ground-View Input}
\label{sec:dataset:platform}
A \dataset scene is one synchronized $360^\circ$ observation from the ground. It comprises four time-synchronized RGB views, front, right, back, and left, from the roof-mounted surround cameras of the GOOSE recording vehicle MuCAR-3~\cite{mortimer2024goose} at $90^\circ$ azimuth spacing, each $2448\times1000$\,px, together with the 128-beam roof-mounted rotating Velodyne LiDAR sweep captured at the same instant: a single point cloud available for every scene, with a median of 200{,}000 points (from 30{,}000 in open terrain to 380{,}000 under dense canopy) and most returns within ${\sim}66$\,m horizontally. We use the surround cameras rather than the windshield camera that GOOSE annotates for semantic segmentation because their full $360^\circ$ coverage matches the top-down geometry of the aerial reference. Releasing both modalities per scene lets \dataset support image-only localization as well as the image-LiDAR fusion architecture we propose.

\subsection{Ground Truth and Aerial Reference}
\label{sec:dataset:gt}
The ground-truth pose of each scene comes from the vehicle's INS
trajectory: we use the RTK-corrected NCom pose (\texttt{pose\_lat}/\texttt{pose\_lon}) of the Oxford RT3000v3 INS where available and fall back to its raw GNSS fix otherwise. The pose is projected from WGS84 to UTM zone~32N (EPSG:25832), Bavaria's native grid. 
Details on the INS localization quality are in the appendix.
 
As the aerial reference we use the Bavarian DOP20 digital orthophotos at their native ground resolution of 20\,cm/px. We cache source orthophotos as $1\,\text{km}\times1\,\text{km}$ tiles ($5000\times5000$\,px) and, for each scene, cut a $1024\times1024$\,px crop ($204.8\,\text{m}\times204.8\ \text{m}$) at native resolution, compositing up to four adjacent source tiles when the crop straddles a tile boundary. Crops are north-aligned, so the image axes coincide with UTM easting and northing. Crucially, the crop is not centered on the vehicle: the ground-truth vehicle pixel is placed uniformly at random within the central $512\times512$ region (within $\pm256$\,px${=}{\pm}51.2$\,m of the tile center, independent of heading), so that the full $204.8$\,m of context is retained and the vehicle is always at least $51.2$\,m from any edge, yet never at a fixed location a method could exploit, following VIGOR's practice of decoupling the query position from the tile center~\cite{zhu2021vigor}. The localization target is this pixel, which must be recovered by cross-view matching. Because the orthophotos are flown on dates independent of the ground drives, every ground-aerial pair carries a built-in temporal domain gap (vegetation state, snow cover, construction), which we treat as a core property of the benchmark rather than noise to be removed. All tiles were flown in June, July, or August and therefore show summer conditions only; 95\% date from 2024 and 5\% from August 2025.
 
\begin{figure}[t]
  \centering
  \includegraphics[width=\linewidth]{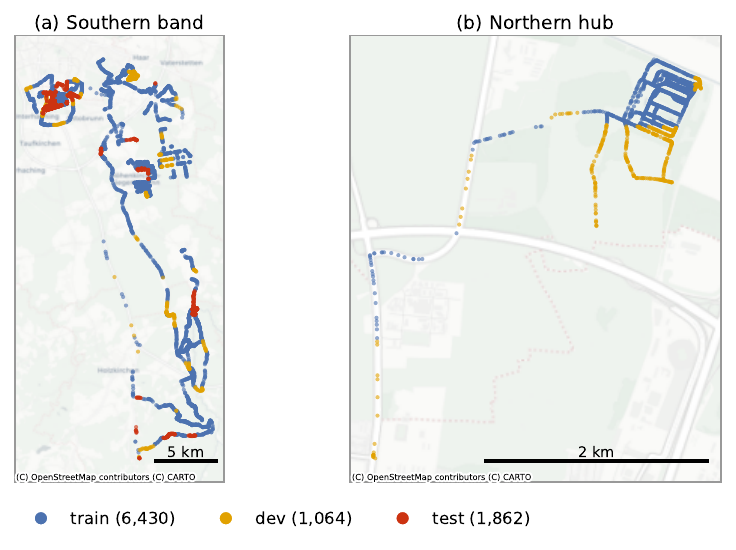}
  \caption{Geographic layout of \dataset across a $\sim$45\,km\,$\times$\,16\,km band of Upper Bavaria south of Munich, shown as (a) the southern band and (b) the northern hub. Scenes are colored by split and train/dev/test occupy geographically partitioned regions. Basemap \copyright\ OpenStreetMap contributors, \copyright\ CARTO.}
  \label{fig:splitmap}
\end{figure}

\dataset is geographically confined to a $\sim$45\,km\,$\times$\,16\,km band of Upper Bavaria south of Munich (latitude $47.81^\circ$--$48.24^\circ$, longitude $11.59^\circ$--$11.81^\circ$), captured between July 2022 and July 2023, as shown in \Cref{fig:splitmap}. Each scene additionally carries a manually verified \textbf{paved}/\textbf{unpaved} terrain label, used both to characterize the dataset and to balance the splits (\cref{sec:dataset:splits}).
\subsection{Composition and Splits}
\label{sec:dataset:splits}
\Cref{tab:splits} summarizes the three splits. The dataset is close to terrain-balanced overall (45.8\% paved, 54.2\% unpaved), and both train and test are individually near-balanced (46.5\% and 48.3\% paved), so the paved/unpaved axis can be studied in isolation. 
\begin{table}[t]
  \centering
  \small
  \setlength{\tabcolsep}{3pt}
  \begin{tabular}{lrrrrrr}
    \toprule
     & \multicolumn{4}{c}{Season} & & \\
    \cmidrule(lr){2-5}
    Split & Winter & Spring & Summer & Autumn & Paved & Total \\
    \midrule
    \texttt{train} & 1{,}387 & 1{,}976 & 1{,}341 & 1{,}726 & 2{,}988 & 6{,}430 \\
    \texttt{dev}   &   154   &   304   &   103   &   503   &   400   & 1{,}064 \\
    \texttt{test}  &   332   &   608   &   532   &   390   &   900   & 1{,}862 \\
    \midrule
    Total          & 1{,}873 & 2{,}888 & 1{,}976 & 2{,}619 & 4{,}288 & 9{,}356 \\
    \bottomrule
  \end{tabular}
  \caption{\textbf{Composition of \dataset.} Scene counts per split by season and
  terrain. Scenes that are not paved are unpaved ($45.8\%$ paved overall).}
  \label{tab:splits}
\end{table}
\textbf{Split Partitioning:} We split the data by place rather than by drive, in the spirit of VIGOR's cross-area protocol~\cite{zhu2021vigor}, which measures generalization to regions unseen during training. Therefore, we overlay a regular grid of $1\,\text{km}\times1\,\text{km}$ cells on the recording area and assign each cell, with every scene inside it, entirely to one split, so that train, dev, and test cover disjoint patches of ground. A model therefore cannot score well simply by recognizing a location it has already seen in training. The only contact between splits is at cell borders, where a drive that crosses into a neighboring cell is cut, with each part kept in its respective set. In the appendix, we show that this border contact has no significant effect on localization performance. Which cell goes to which split is chosen by a search that balances the three splits in size ($\approx$70/10/20), terrain class and season (\Cref{tab:splits}). Finally, the test set keeps only pose-verified scenes, whose GNSS fix and INS pose agree to within $5\,$m, so that test ground-truth positions are reliable.

\begin{figure}
\centering
    \includegraphics[width=\linewidth]{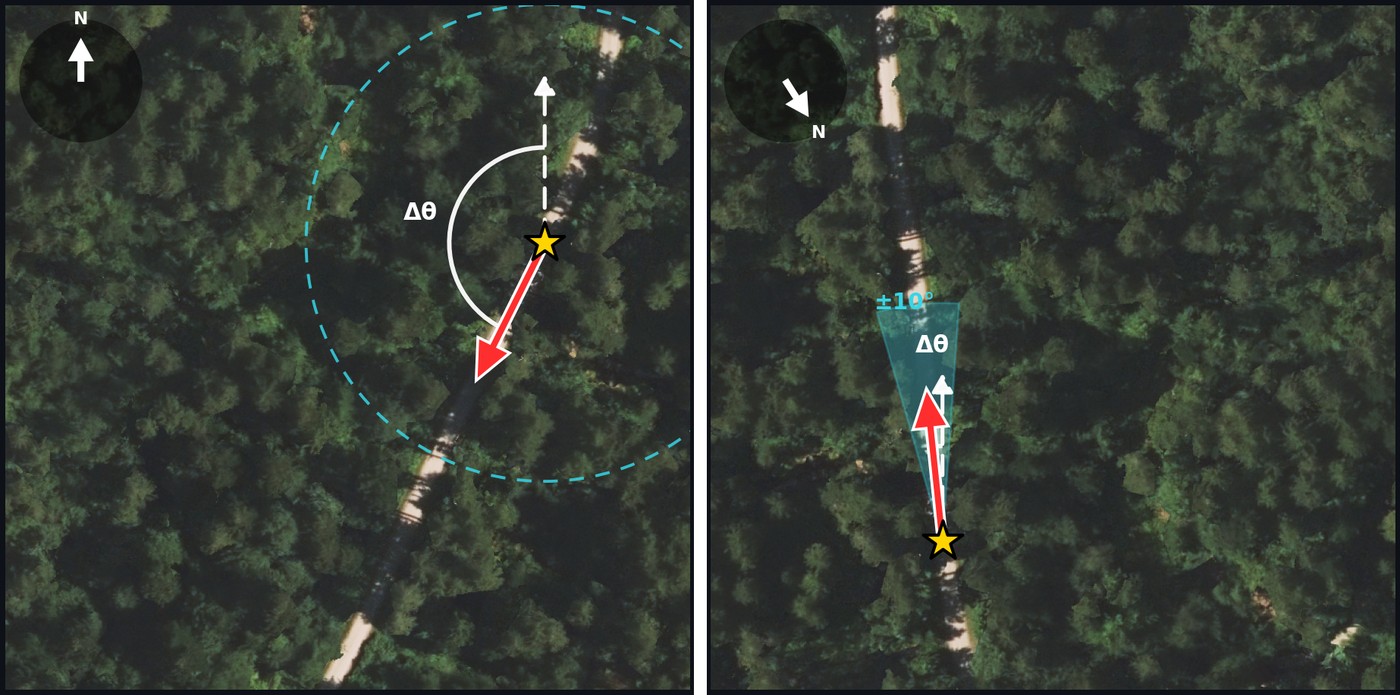}
    \caption{\textbf{The two heading settings} on one scene ($\star$ = GT
    position, red = true heading, white dashed = reference ``up'', cyan =
    admissible range of $\Delta\theta$). Left -- north-up: no prior, recover the
    full $360^\circ$. Right -- IMU prior: $\pm10^\circ$ coarse prior, recover only
    the residual $|\Delta\theta|\le10^\circ$ (offset sampled uniformly in
    $[-10^\circ,+10^\circ]$ at train and test).}
    \label{fig:settings}
\end{figure}

\textbf{Heading settings:} We evaluate under two heading settings that differ only in the heading prior given to the model. The north-up setting provides no orientation information, as the model receives only north-aligned overhead imagery. The IMU-prior setting limits the heading uncertainty to $\pm10^\circ$ by providing overhead imagery that is pre-rotated with a random uniform mismatch of up to $\pm10^\circ$, reflecting the coarse heading available in deployment. North-up therefore serves as the harder stress test. \Cref{fig:settings} illustrates both settings.
\begin{figure*}[t]
    \centering
    \includegraphics[width=0.9\linewidth]{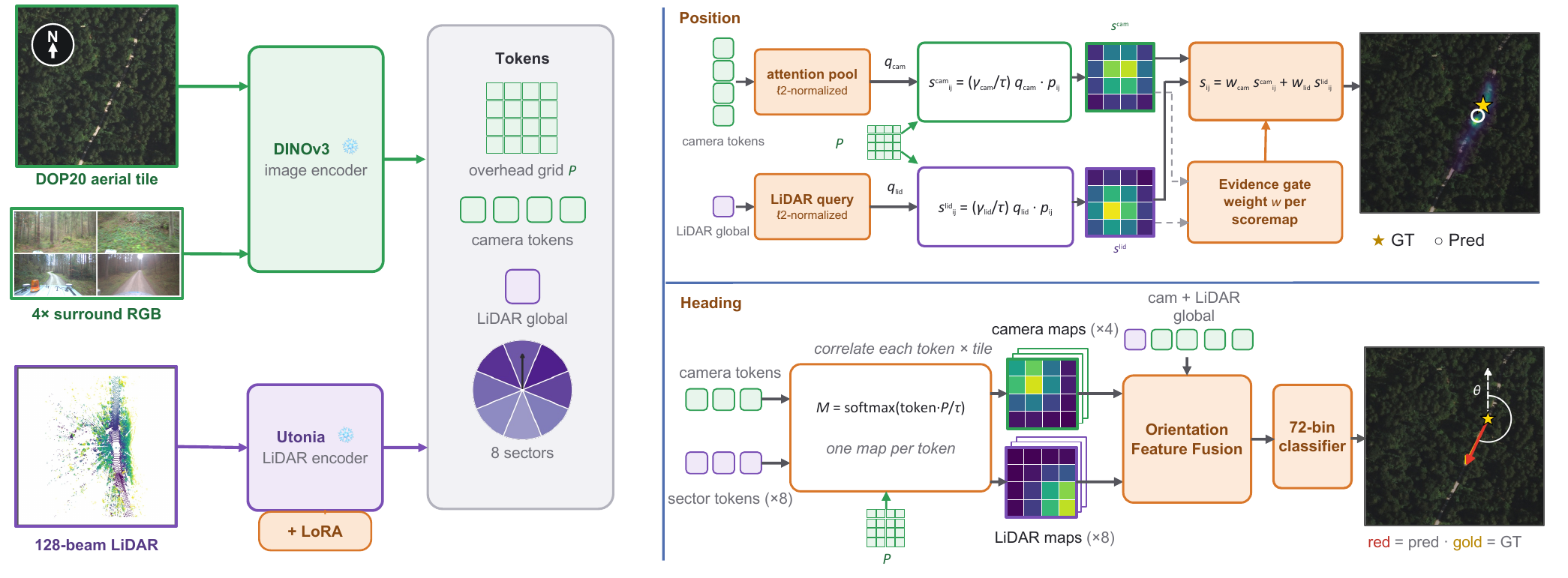}
    \caption{\textbf{Our architecture.} Frozen DINOv3 and Utonia yield a shared token set. In the position branch each modality correlates its own query with the DOP20 patch features, and an evidence gate weights the two resulting score maps before they are summed. The orientation branch correlates the camera and LiDAR-wedge tokens against the tile and feeds the match maps, with the camera [CLS] tokens, to a 72-bin heading classifier.}
    \label{fig:gooselocarch}
\end{figure*}
\section{Methodology}
\label{sec:method}
To establish reference performance on our benchmark, we provide multiple baselines. We first adapt four existing methods to our setting: CCVPE~\cite{xia2023convolutional}, Shi et al.~\cite{shi2023boosting}, HC-Net~\cite{wang2023fine} and FG$^2$~\cite{xia2025fg}. These serve as a competitive reference and show that current methods struggle in off-road environments. Since CCVPE only supports north-aligned input and Shi et al.\ requires a bounded heading prior, we report each method only in the corresponding setting. As CCVPE, Shi et al., HC-Net and FG$^2$ are purely image-based and cannot exploit LiDAR, we add a second, multimodal baseline that fuses image and LiDAR features. This demonstrates that our dataset enables multimodal localization and provides an initial indication of its benefits. Implementation details for reproducing all methods are provided in the appendix.
\subsection{Our multimodal baseline}
We build our multimodal baseline on two frozen foundation encoders: DINOv3~\cite{simeoni2025dinov3} for the image inputs and Utonia~\cite{zhang2026utonia} for the LiDAR point clouds. The two tasks in our benchmark, position and heading prediction, are handled by two separate heads, which are illustrated in \Cref{fig:gooselocarch}. The overhead tile and the four camera views are encoded by a DINOv3 ViT-L/16 that is kept frozen throughout. For the point cloud we likewise keep Utonia frozen and train only a LoRA adapter~\cite{hu2022lora}. This adaptation is necessary because without it the LiDAR branch does not contribute meaningfully, which we attribute to the large domain shift between Utonia's pre-training data and off-road GOOSELoc. We feed the $1024\times1024$ overhead tile to DINOv3, which yields a $64\times64$ patch grid, and a linear layer projects its feature dimension from $1024$ to a $512$-d match space. The four camera views are encoded likewise and their \texttt{[CLS]} tokens are projected to the same $512$-d space. The LiDAR branch produces two features: a mean-pooled global descriptor and $K = 8$ vehicle‑frame sector descriptors, obtained by binning each point by its ego‑frame azimuth $\operatorname{atan2}(y,x)$ into $8$ equal $45.0^\circ$ wedges and mean‑pooling the per‑point Utonia features within each wedge. 
\paragraph{Position Prediction}
\label{sec:method:pos}
We project the overhead patch tokens, the four camera tokens and a global-mean LiDAR token into a shared $d{=}512$ space with a learnable linear projection. The camera tokens additionally receive a learned azimuth embedding so the model can distinguish front from back. Rather than fusing the two modalities into a single query before matching, we let each modality match the tile on its own and fuse the two resulting score maps. A single-query attention pool over the four camera tokens yields a camera query $q_{\mathrm{cam}}$, and the projected LiDAR token is the LiDAR query $q_{\mathrm{lid}}$. Both are $\ell_2$-normalized and correlated with every projected patch $p_{ij}$ of the $64\times64$ grid, giving one score map per modality,
\begin{equation}
  s^{m}_{ij}=\frac{\gamma_{m}}{\tau}\,\hat{q}_{m}^{\top}p_{ij},
  \qquad m\in\{\mathrm{cam},\mathrm{lid}\},
  \label{eq:posmaps}
\end{equation}
with the shared learned temperature $\tau$ (initialized to $0.07$) and a learned per-modality gain $\gamma_{m}$ parameterized in log space and initialized to $\gamma_m{=}1$, which restores the sharpness freedom that the normalization removes.

In our experiments, we observed that the contributions of the LiDAR and RGB ground views vary with the scene. We therefore add an evidence gate that estimates how much each score map $s^m$ should contribute. First we turn each map into a distribution $\pi^{m}=\mathrm{softmax}_{ij}(s^{m})$ which is used only inside the gate. From each $\pi^{m}$ we take three scale-free statistics that say how peaked and how unambiguous the match is: the normalized entropy $-\sum_{ij}\pi^{m}_{ij}\log\pi^{m}_{ij}/\log(64^{2})$, the peak probability $\max_{ij}\pi^{m}_{ij}$, and the gap between the largest and the second-largest value. Stacking these three numbers for both modalities gives a six-dimensional gate input $\beta$, which a two-layer MLP maps to the weights
\begin{equation}
  w=n_{\mathrm{act}}\cdot\mathrm{softmax}\bigl(\mathrm{MLP}(\beta)\bigr)\in\mathbb{R}^{2}.
\end{equation}
The factor $n_{\mathrm{act}}$, the number of active modalities, makes the weights sum to $n_{\mathrm{act}}$, so the scale of the fused logits stays the same when one modality is masked out. The statistics are detached: the gate re-weights the match evidence, but no gradient reaches the maps through it. Based on these weights, the maps are fused as
\begin{equation}
  s_{ij}=w_{\mathrm{cam}}\,s^{\mathrm{cam}}_{ij}
        +w_{\mathrm{lid}}\,s^{\mathrm{lid}}_{ij},
  \qquad
  H=\mathrm{softmax}_{ij}(s).
  \label{eq:posheat}
\end{equation}
The final distribution map $H$ is then used to decode the position by soft-argmax and a KL divergence loss is used on a soft Gaussian target as objective. We use this dense target for two reasons specific to our setting. First, the ground-truth location is continuous and often falls near a patch boundary, where the two adjacent patches are almost equally correct: an InfoNCE~\cite{oord2018representation} target, which is often used in contrastive settings, labels exactly one patch positive and penalizes its neighbor as a hard negative, whereas the soft-Gaussian spreads mass over neighboring patches by distance and barely penalizes a near-boundary prediction. Second, off-road terrain is highly self-similar, so several patches can be genuinely plausible; a one-hot target forces over-confidence against this irreducible ambiguity, while the KL-to-Gaussian objective only asks the model to concentrate mass locally and tolerates a calibrated spread.
\paragraph{Heading Prediction}
Heading is the rotation that aligns the vehicle with the north-aligned tile. This rotation is invisible in appearance alone and it shows up only in where each vehicle-fixed sensor matches the tile. We reuse the projected patch tokens $p_{ij}$, here with a stop-gradient, and write $c_c$ and $u_k$ for the projected camera and LiDAR-wedge tokens. For each surround camera we correlate its token with every patch and soft-max over the $64\times64$ grid to form a cross-view match map,
\begin{equation}
M^{\text{cam}}_{c}[i,j]=\operatorname*{softmax}_{ij}
\frac{\langle c_c,\,p_{ij}\rangle}{\tau},
\qquad c=1,\dots,4 .
\label{eq:cammap}
\end{equation}
Each camera points along a fixed vehicle bearing, so its match map peaks where that bearing meets the tile: ahead for the front camera, behind for the rear, and at $\pm 90^\circ$ for the two side cameras. The head reads the vehicle's heading off the angle of this four-peak pattern, a generalized Hough vote over orientations that is learned rather than hand-designed. Because the four bearings are observed jointly, it also resolves the front/back ($\theta$ vs.\ $\theta{+}180^\circ$) ambiguity that a single forward camera cannot.

We stack the four maps as a four-channel image and pass them through a small two-layer CNN, $\phi_{\text{cam}}=\mathrm{CNN}_{\text{cam}}(M^{\text{cam}}_{1:4}) \in\mathbb{R}^{512}$. The strided convolutions keep a coarse $4\times4$ grid, so the descriptor retains where each peak lies. A global pool would discard precisely the location that carries the rotation. We treat the $K{=}8$ vehicle-frame LiDAR wedges the same way: each pooled wedge is correlated against the patches, $M^{\text{lid}}_{k}[i,j]=\operatorname*{softmax}_{ij}\langle u_k,p_{ij}\rangle/\tau$, and the $K$ maps are reduced by a second two-layer CNN to $\phi_{\text{lid}}\in\mathbb{R}^{512}$. A raw wedge descriptor is invariant to the tile's orientation; only its match location moves as the vehicle turns, which is what makes it informative here.
Two weak appearance cues complete the input. We keep the raw camera \texttt{[CLS]} tokens $z_{1:4}$, and add an overhead-conditioned cross-attention read-out in which each camera token attends over the raw overhead tokens $o$ under a fixed 2D positional encoding $\boldsymbol{\rho}$,
\begin{equation}
  a_c=\operatorname{Attn}\!\big(W_q z_c,\;\; W_k o+\boldsymbol{\rho},
  \;\; W_v o+\boldsymbol{\rho}\big)\in\mathbb{R}^{d}.
\label{eq:xattn}
\end{equation}
\begin{table*}[t]
    \centering
    \small
      \setlength{\tabcolsep}{4pt}
      \begin{tabular}{ll cc cc cc cc}
          \toprule
          & & \multicolumn{4}{c}{dev} & \multicolumn{4}{c}{test} \\
          \cmidrule(lr){3-6}\cmidrule(lr){7-10}
          & & \multicolumn{2}{c}{$\downarrow$\,Localization (m)} & \multicolumn{2}{c}{$\downarrow$\,Orientation
  ($^\circ$)}
              & \multicolumn{2}{c}{$\downarrow$\,Localization (m)} & \multicolumn{2}{c}{$\downarrow$\,Orientation
  ($^\circ$)} \\
          \cmidrule(lr){3-4}\cmidrule(lr){5-6}\cmidrule(lr){7-8}\cmidrule(lr){9-10}
          Setting & Methods & Mean & Median & Mean & Median & Mean & Median & Mean & Median \\
          \midrule
          North-up & Predict center    & 38.9 & 40.9 & 90.0 & 90.0 & 39.7 & 41.5 & 90.0 & 90.0 \\
                   & CCVPE (surround)  & 33.6{\scriptsize$\pm$0.4} & 25.7{\scriptsize$\pm$1.3} &
  37.2{\scriptsize$\pm$8.5} & 16.7{\scriptsize$\pm$12.2} & 45.4{\scriptsize$\pm$2.9} & 37.0{\scriptsize$\pm$2.9} &
  52.8{\scriptsize$\pm$5.0} & 35.6{\scriptsize$\pm$17.0} \\
                   & HC-Net (surround) & 31.3{\scriptsize$\pm$0.2} & 27.3{\scriptsize$\pm$0.6} &
  85.7{\scriptsize$\pm$1.5} & 84.2{\scriptsize$\pm$4.8} & 31.8{\scriptsize$\pm$0.4} & 28.6{\scriptsize$\pm$0.2} &
  87.0{\scriptsize$\pm$0.5} & 84.4{\scriptsize$\pm$1.2} \\
                   & FG$^2$ (front)    & 32.8{\scriptsize$\pm$0.9} & 29.3{\scriptsize$\pm$0.2} &
  89.9{\scriptsize$\pm$0.4} & 91.3{\scriptsize$\pm$0.4} & 37.7{\scriptsize$\pm$1.2} & 35.1{\scriptsize$\pm$0.8} &
  92.0{\scriptsize$\pm$0.2} & 93.4{\scriptsize$\pm$0.1} \\
                   & FG$^2$ (surround) & 30.4{\scriptsize$\pm$0.2} & 28.0{\scriptsize$\pm$0.3} &
  89.7{\scriptsize$\pm$0.1} & 90.4{\scriptsize$\pm$0.1} & 34.4{\scriptsize$\pm$0.4} & 32.1{\scriptsize$\pm$0.3} &
  92.5{\scriptsize$\pm$0.1} & 93.1{\scriptsize$\pm$0.3} \\
                   & Ours (RGB)        & 25.9{\scriptsize$\pm$0.1} & 21.8{\scriptsize$\pm$0.2} &
  30.2{\scriptsize$\pm$4.0} & 11.5{\scriptsize$\pm$0.9} & 29.8{\scriptsize$\pm$0.1} & 25.5{\scriptsize$\pm$0.1} &
  49.0{\scriptsize$\pm$5.3} & 27.9{\scriptsize$\pm$3.3} \\
                   & Ours (LiDAR-only) & 23.9{\scriptsize$\pm$0.4} & 19.8{\scriptsize$\pm$0.4} &
  58.3{\scriptsize$\pm$4.0} & 16.5{\scriptsize$\pm$2.0} & 28.7{\scriptsize$\pm$0.3} & 24.9{\scriptsize$\pm$0.8} &
  64.3{\scriptsize$\pm$3.6} & 38.5{\scriptsize$\pm$5.4} \\
                   & Ours (RGB+L)      & \textbf{23.1{\scriptsize$\pm$0.3}} & \textbf{18.8{\scriptsize$\pm$0.2}} &
  \textbf{26.2{\scriptsize$\pm$1.0}} & \textbf{10.2{\scriptsize$\pm$0.4}} & \textbf{27.1{\scriptsize$\pm$0.3}} &
  \textbf{22.2{\scriptsize$\pm$0.6}} & \textbf{46.3{\scriptsize$\pm$1.1}} & \textbf{24.8{\scriptsize$\pm$0.3}} \\
          \midrule
          IMU $\pm10^\circ$ & Shi et al.\ (front) & 32.4 & 22.7 & \textbf{3.3} & \textbf{2.5} & 35.4 & 26.6 &
  \textbf{4.1} & \textbf{3.2} \\
                   & HC-Net (surround) & 29.8{\scriptsize$\pm$0.5} & 26.2{\scriptsize$\pm$0.5} &
  12.4{\scriptsize$\pm$0.6} & 8.9{\scriptsize$\pm$0.5} & 31.9{\scriptsize$\pm$3.3} & 27.2{\scriptsize$\pm$0.3} &
  15.0{\scriptsize$\pm$1.5} & 11.7{\scriptsize$\pm$0.8} \\
                   & FG$^2$ (front)    & 23.1{\scriptsize$\pm$0.4} & 17.0{\scriptsize$\pm$0.4} &
  5.2{\scriptsize$\pm$0.2} & 3.7{\scriptsize$\pm$0.1} & 29.2{\scriptsize$\pm$0.3} & 24.6{\scriptsize$\pm$0.6} &
  6.6{\scriptsize$\pm$0.1} & 4.8{\scriptsize$\pm$0.1} \\
                   & FG$^2$ (surround) & 19.8{\scriptsize$\pm$0.2} & 14.5{\scriptsize$\pm$0.3} &
  4.3{\scriptsize$\pm$0.1} & 3.6{\scriptsize$\pm$0.1} & 25.0{\scriptsize$\pm$0.4} & 20.2{\scriptsize$\pm$0.6} &
  5.1{\scriptsize$\pm$0.1} & 4.4{\scriptsize$\pm$0.2} \\
                   & Ours (RGB)        & 21.1{\scriptsize$\pm$0.0} & 16.3{\scriptsize$\pm$0.1} &
  3.8{\scriptsize$\pm$0.1} & 3.0{\scriptsize$\pm$0.1} & 25.1{\scriptsize$\pm$0.1} & 20.7{\scriptsize$\pm$0.2} &
  4.6{\scriptsize$\pm$0.1} & 3.9{\scriptsize$\pm$0.1} \\
                   & Ours (LiDAR-only) & 16.9{\scriptsize$\pm$0.2} & 11.3{\scriptsize$\pm$0.2} &
  3.4{\scriptsize$\pm$0.0} & 2.6{\scriptsize$\pm$0.1} & 21.2{\scriptsize$\pm$0.2} & 16.1{\scriptsize$\pm$0.1} &
  4.2{\scriptsize$\pm$0.1} & 3.5{\scriptsize$\pm$0.1} \\
                   & Ours (RGB+L)      & \textbf{16.2{\scriptsize$\pm$0.2}} & \textbf{10.5{\scriptsize$\pm$0.3}} &
  3.7{\scriptsize$\pm$0.1} & 2.8{\scriptsize$\pm$0.0} & \textbf{19.5{\scriptsize$\pm$0.3}} &
  \textbf{14.1{\scriptsize$\pm$0.1}} & 4.6{\scriptsize$\pm$0.1} & 3.8{\scriptsize$\pm$0.1} \\
          \bottomrule
        \end{tabular}%
        \caption{\textbf{Baseline localization on \dataset{} (dev and test).} Mean and median error for position (m) and orientation ($^\circ$), averaged over three seeds. Predict center is a non-learned trivial baseline. We evaluate FG$^2$ on the native front camera and on our surround-view adaptation, CCVPE and HC-Net on our surround pseudo-panorama, and Shi et al. on the front camera (its native configuration). CCVPE and HC-Net appear only in the north-aligned setting, Shi et al.\ only under the IMU prior (bounded-prior method). Best per column within each setting in \textbf{bold}. The IMU-prior setting predicts a bounded residual $|\Delta\theta|\le10^\circ$.}
    \label{tab:results}
\end{table*}
The positional encoding makes this read-out carry the match location, so it acts as a soft, learned counterpart of the explicit maps. The heading feature concatenates these blocks,
\begin{equation}
f=\big[\,z_{1:4}\,;\; a_{1:4}\,;\; \phi_{\text{cam}}\,;\; \phi_{\text{lid}}\,\big],
\end{equation}
and we decode heading with a $72$-bin ($5^\circ$) circular classifier. The classifier can place mass on both $\theta$ and $\theta{+}180^\circ$ and commit to the better-supported mode, which is what the front/back symmetry requires. At test time we take the arg-max bin refined by a circular centroid over its two neighbors to obtain sub-$5^\circ$ resolution.
\paragraph{Training Objective}
Both heads are trained jointly with a single composite objective,
\begin{equation}
\mathcal{L} \;=\; \underbrace{\mathrm{KL}\!\big(g_\sigma \,\Vert\, H\big)}_{\text{position}}
\;-\; \underbrace{\sum_{k=1}^{72}\tilde{g}_k\log \hat{y}_k}_{\text{heading}},
\label{eq:loss}
\end{equation}
where the position term is the KL divergence from the soft-Gaussian target $g_\sigma$ ($\sigma{=}1.5$ patches) to the predicted heat-map $H$, and the heading term is a soft circular cross-entropy between the $72$-bin prediction $\hat{y}$ and a wrapped-Gaussian soft label $\tilde{g}$ ($\sigma{=}10^\circ$) centered on the ground-truth heading, which smooths the hard bin labels and respects the circular topology.
\section{Experiments}
\label{sec:experiments}
We compare our approach with CCVPE, Shi et al., HC-Net, FG$^2$ and a trivial center-prediction baseline in \Cref{tab:results}. In the north-up setting all three variants of our approach outperform every baseline in localization on both dev and test, and the camera-bearing variants (RGB, RGB+L) additionally lead in orientation. The picture is more nuanced under the IMU prior: the fusion again yields the best localization, while Shi et al.\ attains slightly better orientation, at the cost of the weakest mean localization in this setting. Our RGB variant is slightly weaker than FG$^2$ (surround) on dev but generalizes to test on par with it. It is worth noting that FG$^2$ sits near $90^\circ$ mean orientation error in the north-up setting, i.e.\ chance for a $360^\circ$ heading.

LiDAR and cameras play complementary roles. LiDAR is the stronger single modality for localization: 3D structure is a strong position cue against the overhead map. Without a heading prior, however, it cannot orient, since road and forest geometry are largely front/back symmetric; the cameras contribute the appearance information that breaks this $180^\circ$ ambiguity. The fusion inherits both strengths and improves on each: it is more accurate than LiDAR-only in every localization column (paired per-scene difference $-1.7$\,m on test under the IMU prior) while matching the camera-level heading. The remaining north-up orientation error is heavy-tailed: a few near-$180^\circ$ front/back flips, most frequent for LiDAR-only, pull the mean far above the median, so typical scenes are predicted far better than the mean suggests.
  
The heading prior helps both tasks: it collapses orientation error to a few degrees and, by resolving the tile rotation, also lowers localization error. Absolute errors nonetheless remain large throughout, leaving clear headroom on \dataset{} and supporting our framing of these models as reference baselines.

\subsection{Ablation}
We label each test scene as paved or unpaved and tag it by season, and report median test error per terrain (\cref{tab:terrain}) and season (\cref{tab:season}); further architectural ablations are in the appendix.

\begin{table}[t]
        \centering
        \resizebox{\linewidth}{!}{
        \begin{tabular}{ll cc cc}
          \toprule
          & & \multicolumn{2}{c}{Paved} & \multicolumn{2}{c}{Unpaved} \\
          \cmidrule(lr){3-4}\cmidrule(lr){5-6}
          Setting & Method & Loc\,(m) & Ori\,($^\circ$) & Loc\,(m) & Ori\,($^\circ$) \\
          \midrule
          North-up & FG$^2$ (surround) & 27.6{\scriptsize$\pm$0.4} & 89.9{\scriptsize$\pm$0.2} &
  36.1{\scriptsize$\pm$0.4} & 98.0{\scriptsize$\pm$0.9} \\
                   & Ours (RGB)        & 25.5{\scriptsize$\pm$0.7} & 28.6{\scriptsize$\pm$3.3} &
  25.5{\scriptsize$\pm$0.3} & 27.6{\scriptsize$\pm$4.2} \\
                   & Ours (LiDAR-only) & 23.2{\scriptsize$\pm$0.6} & 35.6{\scriptsize$\pm$7.3} &
  26.5{\scriptsize$\pm$0.9} & 40.8{\scriptsize$\pm$3.9} \\
                   & Ours (RGB+L)      & \textbf{21.2{\scriptsize$\pm$0.3}} & \textbf{24.8{\scriptsize$\pm$1.6}} &
  \textbf{22.9{\scriptsize$\pm$0.7}} & \textbf{25.3{\scriptsize$\pm$1.1}} \\
          \midrule
          IMU $\pm10^\circ$ & FG$^2$ (surround) & 13.8{\scriptsize$\pm$0.1} & 4.0{\scriptsize$\pm$0.2} &
  27.0{\scriptsize$\pm$0.7} & 4.7{\scriptsize$\pm$0.2} \\
                   & Ours (RGB)        & 19.5{\scriptsize$\pm$0.5} & 3.9{\scriptsize$\pm$0.1} &
  21.5{\scriptsize$\pm$0.2} & 3.8{\scriptsize$\pm$0.0} \\
                   & Ours (LiDAR-only) & 13.4{\scriptsize$\pm$0.1} & \textbf{3.5{\scriptsize$\pm$0.2}} &
  19.1{\scriptsize$\pm$0.5} & \textbf{3.5{\scriptsize$\pm$0.1}} \\
                   & Ours (RGB+L)      & \textbf{12.2{\scriptsize$\pm$0.3}} & 3.8{\scriptsize$\pm$0.1} &
  \textbf{15.4{\scriptsize$\pm$0.3}} & 3.8{\scriptsize$\pm$0.1} \\
          \bottomrule
        \end{tabular}
        }
        \caption{\textbf{Performance by terrain (test, median $\pm$ std over 3 seeds).}}
        \label{tab:terrain}
\end{table}
\textbf{Terrain:} FG$^2$ is competitive only on paved roads with a prior, its native KITTI regime, and its error roughly doubles off-road for the IMU-prior setting, whereas our fusion degrades far less and is best in the off-road regime \dataset{} targets. For orientation, the camera variants are essentially terrain-invariant, while LiDAR-only degrades off-road; the fusion keeps camera-level orientation on both terrains.
\begin{table}[t]
        \centering
        \resizebox{\linewidth}{!}{
        \begin{tabular}{ll cc cc cc cc}
          \toprule
          \multicolumn{2}{c}{Setting} & \multicolumn{2}{c}{Winter} & \multicolumn{2}{c}{Spring} &
  \multicolumn{2}{c}{Summer} & \multicolumn{2}{c}{Autumn} \\
          \cmidrule(lr){3-4}\cmidrule(lr){5-6}\cmidrule(lr){7-8}\cmidrule(lr){9-10}
          \multicolumn{2}{c}{/ Method} & Loc(m) & Ori($^\circ$) & Loc(m) & Ori($^\circ$) & Loc(m) & Ori($^\circ$) &
  Loc(m) & Ori($^\circ$) \\
          \midrule
          North-up  & FG$^2$ & 28.4{\scriptsize$\pm$1.2} & 80.3{\scriptsize$\pm$0.8} & 28.5{\scriptsize$\pm$0.7} &
  90.2{\scriptsize$\pm$0.9} & 36.5{\scriptsize$\pm$0.6} & 100.1{\scriptsize$\pm$1.5} & 34.6{\scriptsize$\pm$0.4} &
  108.5{\scriptsize$\pm$0.6} \\
              & RGB    & 25.5{\scriptsize$\pm$0.7} & 26.9{\scriptsize$\pm$2.9} & 24.3{\scriptsize$\pm$0.1} &
  36.5{\scriptsize$\pm$10.2} & 26.3{\scriptsize$\pm$0.6} & \textbf{28.3{\scriptsize$\pm$1.2}} &
  25.9{\scriptsize$\pm$0.2} & 19.5{\scriptsize$\pm$2.7} \\
              & L-only & 23.9{\scriptsize$\pm$1.1} & 44.4{\scriptsize$\pm$4.4} & 23.5{\scriptsize$\pm$1.2} &
  \textbf{25.7{\scriptsize$\pm$4.8}} & 26.1{\scriptsize$\pm$0.7} & 44.8{\scriptsize$\pm$1.4} &
  26.7{\scriptsize$\pm$1.5} & 45.3{\scriptsize$\pm$15.1} \\
              & RGB+L  & \textbf{22.0{\scriptsize$\pm$1.0}} & \textbf{25.3{\scriptsize$\pm$0.7}} &
  \textbf{20.3{\scriptsize$\pm$0.5}} & 30.8{\scriptsize$\pm$0.1} & \textbf{24.1{\scriptsize$\pm$0.5}} &
  28.8{\scriptsize$\pm$2.3} & \textbf{22.7{\scriptsize$\pm$0.8}} & \textbf{16.3{\scriptsize$\pm$1.4}} \\
          \midrule
          IMU & FG$^2$ & 16.9{\scriptsize$\pm$0.6} & 4.6{\scriptsize$\pm$0.1} & 14.2{\scriptsize$\pm$0.1} &
  4.0{\scriptsize$\pm$0.2} & 26.6{\scriptsize$\pm$1.1} & 4.8{\scriptsize$\pm$0.2} & 25.1{\scriptsize$\pm$1.0} &
  4.5{\scriptsize$\pm$0.3} \\
              & RGB    & 19.0{\scriptsize$\pm$0.5} & 4.3{\scriptsize$\pm$0.1} & 18.4{\scriptsize$\pm$0.7} &
  3.6{\scriptsize$\pm$0.2} & 22.8{\scriptsize$\pm$0.1} & 3.7{\scriptsize$\pm$0.1} & 23.5{\scriptsize$\pm$0.3} &
  \textbf{4.0{\scriptsize$\pm$0.2}} \\
              & L-only & 16.8{\scriptsize$\pm$0.2} & \textbf{4.0{\scriptsize$\pm$0.2}} & 13.1{\scriptsize$\pm$0.4} &
  \textbf{3.3{\scriptsize$\pm$0.1}} & 17.1{\scriptsize$\pm$0.1} & \textbf{3.3{\scriptsize$\pm$0.1}} &
  19.5{\scriptsize$\pm$0.7} & \textbf{4.0{\scriptsize$\pm$0.1}} \\
              & RGB+L  & \textbf{13.4{\scriptsize$\pm$0.7}} & 4.2{\scriptsize$\pm$0.3} &
  \textbf{11.7{\scriptsize$\pm$0.0}} & \textbf{3.3{\scriptsize$\pm$0.1}} & \textbf{15.0{\scriptsize$\pm$0.1}} &
  3.8{\scriptsize$\pm$0.1} & \textbf{17.3{\scriptsize$\pm$0.3}} & 4.1{\scriptsize$\pm$0.2} \\
          \bottomrule
        \end{tabular}
        }
        \caption{\textbf{Performance by season (test, median $\pm$ std over 3 seeds).}}
        \label{tab:season}
\end{table}

\textbf{Season:} Localization is hardest in the leaf-on seasons, where dense canopy occludes the overhead view. The fusion localizes best in every season and setting. Under the IMU prior, the LiDAR advantage over RGB grows from winter to summer, consistent with geometry being largely invariant to foliage. Orientation follows a different pattern: camera orientation is weakest in spring and strongest in autumn, LiDAR-only shows the mirror image, and the fusion is best in winter and autumn. FG$^2$ remains near chance throughout.

\begin{figure}[t]
    \centering
    \includegraphics[width=\linewidth]{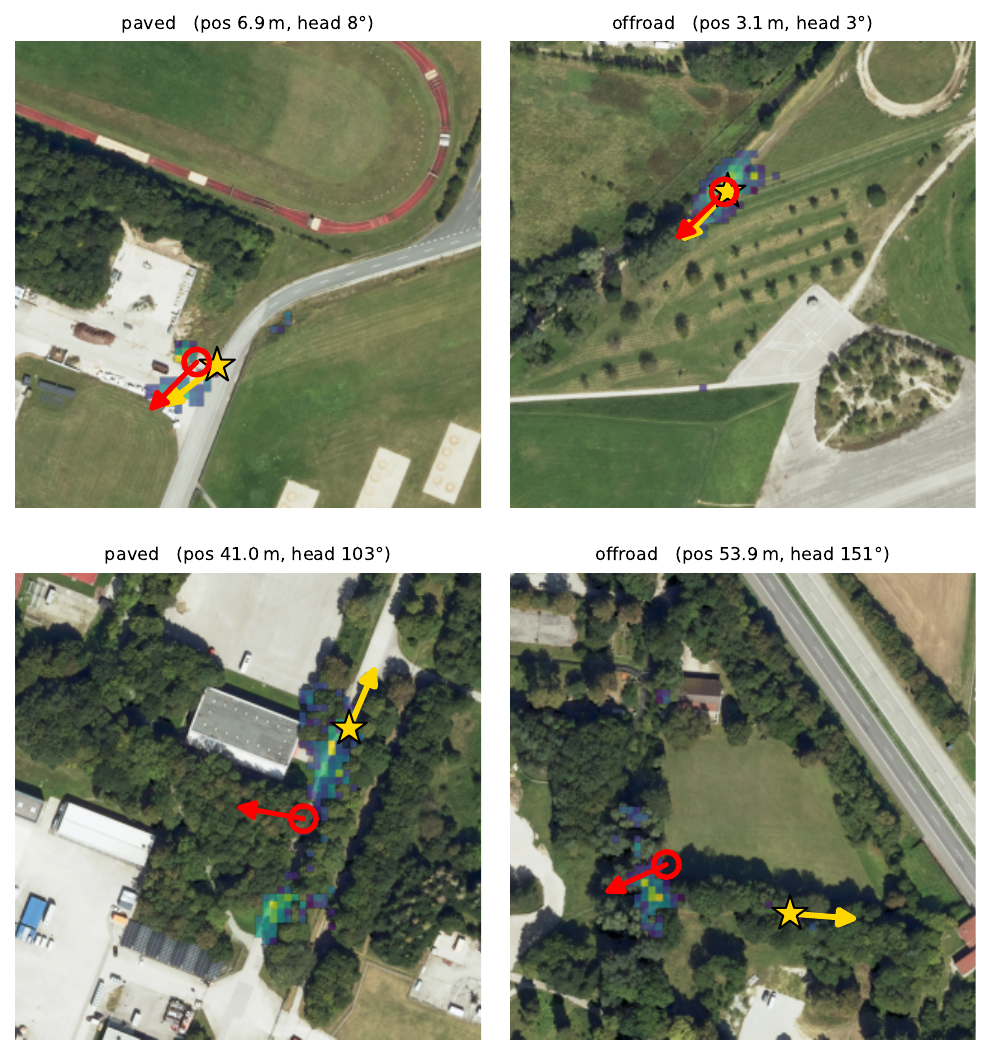}
    \caption{\textbf{Qualitative predictions on \dataset{}} (test, north-up, no prior). Gold $\star$ / arrow: ground-truth position / heading, red $\circ$/arrow: prediction; the viridis overlay is the predicted position posterior.}
    \label{fig:qualitative}
\end{figure}

\textbf{Qualitative Results:} \Cref{fig:qualitative} shows typical successes and the two failure modes: a ${\sim}180^\circ$ front/back flip, and a diffuse posterior whose soft-argmax drifts from the true position.
\section{Conclusion}
We introduced \dataset{}, a multimodal benchmark for metric ground-to-aerial localization spanning the structured-to-unstructured spectrum from paved roads to off-road forest and field tracks. Fine-tuning four state-of-the-art image-only methods reveals that the accuracy achieved on on-road benchmarks decreases significantly off-road and worsens further without a heading prior. Our multimodal baseline narrows this gap and reveals a clear division of labor between the modalities. LiDAR geometry is the stronger position cue and provides the best localization even on its own. In contrast, camera appearance resolves absolute heading. Fusing the two modalities retains both capabilities.

These results also reveal open research questions. Absolute errors remain at tens of meters and tens of degrees without a prior, far from deployment-grade, indicating that cross-view localization in unstructured terrain is a broader issue rather than a failing of any single method. \dataset{} is also deliberately narrow: a single calibrated sensor rig in one region of Upper Bavaria, matched against summer-only orthophotos. Fixing sensor, geography, and aerial source isolates the terrain and seasonal axes we study, but limits external validity: our results characterize this regime, and geographic or sensor generalization must be established on data from other regions and platforms. Broadening this coverage and incorporating multi-frame reasoning are natural next steps. We release the data, splits, and baselines, and hope \dataset{} serves as a reference point for cross-view localization beyond roads.
\newpage
\begin{acks}
The authors gratefully acknowledge the Institute for Distributed Intelligent Systems for granting the computing time and providing access to the Monacum One GPU cluster at the University of the Bundeswehr Munich.

This work was supported by the Federal Office of Bundeswehr Equipment, Information Technology and In-Service Support (BAAINBw).

This work is funded by the Deutsche Forschungsgemeinschaft (DFG, German Research Foundation) under grant number 583569863.
\end{acks}
{
    \small
    \bibliographystyle{ieeenat_fullname}
    \bibliography{main}
}

\clearpage
\appendix
\setcounter{figure}{0}\renewcommand{\thefigure}{A\arabic{figure}}
\setcounter{table}{0}\renewcommand{\thetable}{A\arabic{table}}
\setcounter{equation}{0}\renewcommand{\theequation}{A\arabic{equation}}
\twocolumn[{\centering\Large\bfseries Appendix\par\vspace{1.5em}}]
\section{GOOSELoc Pose Quality}
The Oxford RT3000v3 INS system provides status codes that we can use to quantify the expected error in the ground truth poses used in the different splits of GOOSELoc.\footnote{Visit the OxTS GitHub repository on their NCOMdecoder for more details on the status names: \url{https://github.com/OxfordTechnicalSolutions/NCOMdecoder}.}
These status codes are determined based on common indicators like the number of satellites visible for the GPS receiver INS system, the geometric quality of the satellites' position in the sky and the availability of ground network based GNSS correction data via NTRIP.

For the status code \texttt{RTK Integer} one can expect \si{\centi\meter}-level positioning and from status code \texttt{RTK Float} \si{\deci\meter}-level positioning.
We ensured that for the splits train and test we maintain a close balance based on these INS status codes (\cref{tab:gnss_mode_splits}).

\begin{table}[htpb!]
    \centering
    \setlength{\tabcolsep}{4pt}
    \small
    \begin{tabular}{@{}crccc@{}}
      \toprule
      \multirow{2}{*}{Split} & \multirow{2}{*}{Scenes} &
      \multicolumn{3}{c@{}}{GNSS position mode (\%)} \\
      \cmidrule(l){3-5}
      & & RTK Integer (cm) & RTK Float (dm) & Other \\
      \midrule
      train & 6430 & 74.4          & 10.4 & 15.2 \\
      dev   & 1064 & 89.6 & 5.9  & 4.5 \\
      test  & 1862 & 70.4          & 9.0  & 20.6 \\
      \bottomrule
    \end{tabular}
    \caption{Ground truth pose quality per split, decoded from the
    \texttt{gps\_pos\_mode} byte of the NCOM packets; rows sum to $100\%$.
    ``Other'' groups the status codes \texttt{Differential}, \texttt{SPS} and packets with no position solution at
    all. Pose quality was not a split constraint, but care was taken to maintain a good balance for the train and test split.}
    \label{tab:gnss_mode_splits}
\end{table}
\begin{figure}[htpb!]
    \vspace{2em} 
    \centering
    \begin{tikzpicture}
        \node[anchor=south west,inner sep=0] (img) at (0,0)
        {\includegraphics[width=.96\linewidth, trim=1pt 1pt 1pt 1pt, clip]{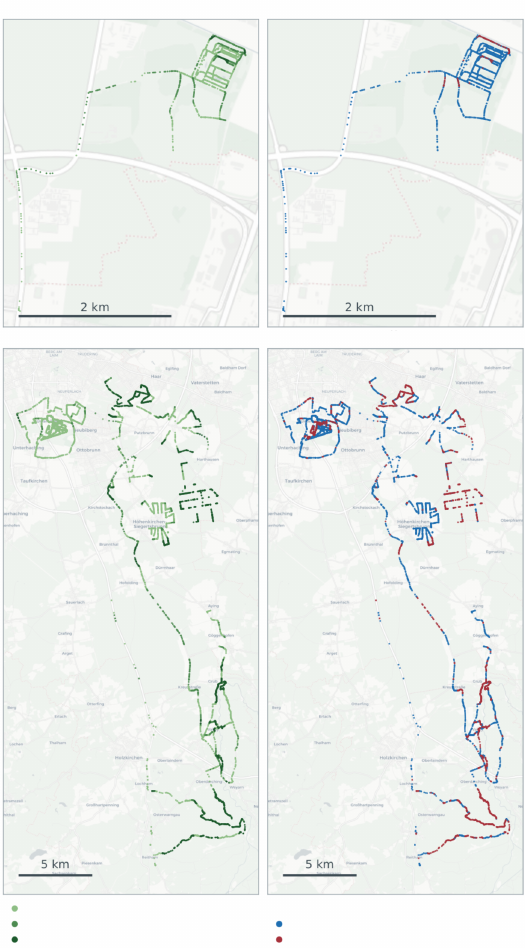}};
        \begin{scope}[x={(img.south east)}, y={(img.north west)}]
            \node[anchor=west, fill=white, draw=gray, inner sep=2pt, text opacity=1] at (0.15,0.983) {\scriptsize Tree Cover};
            \node[anchor=west, fill=white, draw=gray, inner sep=2pt, text opacity=1] at (0.6,0.985) {\scriptsize GNSS Heading Solution};
            \node[anchor=west, fill=white, draw=gray, inner sep=2pt, text opacity=1] at (0.15,0.635) {\scriptsize Tree Cover};
            \node[anchor=west, fill=white, draw=gray, inner sep=2pt, text opacity=1] at (0.6,0.635) {\scriptsize GNSS Heading Solution};
            
            \node[anchor=west,text opacity=1] at (0.03,0.042) {\scriptsize Open Sky};
            \node[anchor=west,text opacity=1] at (0.03,0.025) {\scriptsize Within \SI{25}{\meter} of Forest};
            \node[anchor=west,text opacity=1] at (0.03,0.007) {\scriptsize Inside Forest / Wood};
            \node[anchor=west,text opacity=1] at (0.535,0.025) {\scriptsize RTK Integer / Float Heading};
            \node[anchor=west,text opacity=1] at (0.535,0.007) {\scriptsize Lower Status Codes};
        \end{scope}
  \end{tikzpicture}
  \caption{Geographic layout of GOOSELoc in terms of tree cover and the GNSS heading solution. Basemap \copyright\ OpenStreetMap contributors, \copyright\ CARTO.}
  \label{gooseloc_pose_quality_map}
\end{figure}

Simply expecting the scenes in category ``Other'' to have a much poorer ground-truth pose misrepresents the capability of the INS system. 
The modes \texttt{RTK Integer} and \texttt{RTK Float} can effectively only be achieved under open sky conditions, where multiple satellites have a direct line of sight to the dual antenna setup on the roof of MuCAR-3.
This means that any scenes recorded in or along a forested area suffer from the canopy effect in their GNSS positioning.
This is reflected in the scene statistics, with only 39.7\% of the GOOSELoc scenes recorded within an OpenStreetMap tile marked as \texttt{landuse=forest} having an \texttt{RTK Integer/Float} status.
In comparison, 85.6\% of all scenes more than \SI{25}{\meter} away from a forest have an \texttt{RTK Integer/Float} status (\cref{gooseloc_pose_quality_map}).

The Oxford RT3000v3 INS system uses a high-grade IMU system to maintain accurate localization even under harsh INS conditions. 
The routes with many scenes with lower GNSS heading solutions in GOOSELoc do not contain many turns and contain scenes with RTK accuracy in between, allowing the INS system to keep the accumulated errors from an IMU-only localization minimal.

When manually inspecting the scenes with lower status codes in each split, we could not observe any mismatches.
A human visual inspection without adequate tooling cannot recognize position mismatches at \si{\deci\meter}-scale.
The fact that the best-performing models under ideal open sky conditions like in the category ``Paved'' currently have a median localization error larger than \SI{10}{\meter} gives us confidence that the accuracy of the ground truth data with this type of INS system is suitable as training signal and as a test benchmark.

\section{FG$^2$ baseline}
FG$^2$~\cite{xia2025fg} ingests a single forward-facing ground image, so we apply it to each of the four surround cameras independently rather than fusing them into a single representation. For camera $c$ with fixed azimuth offset $\delta_c \in \{0^\circ,90^\circ,180^\circ,270^\circ\}$, we rotate the aerial tile by $\delta_c$ so that camera $c$ acts as a virtual forward view, and run the unmodified FG$^2$ on the pair (image$_c$, rotated tile). At training time we sample one camera per example, which acts as a $4\times$ orientation augmentation. At inference we run all four cameras, map each predicted pose back into the common tile frame, and fuse them by a confidence-weighted average. Each camera's confidence is the total cross-view matching probability of its weighted-Procrustes solution. The FG$^2$ network itself is left unchanged; for a detailed description
  of the approach we refer to the original paper.
\section{CCVPE baseline}
CCVPE~\cite{xia2023convolutional} is a panoramic cross-view method: it consumes a single $360^\circ$ ground panorama together with a north-up aerial patch covering the query's neighborhood, and jointly predicts a dense location-probability map over the patch and the camera orientation, the latter obtained by matching the ground descriptor against rolled (rotated) copies of the aerial descriptor. Because CCVPE ingests one panoramic image, we fuse the four surround cameras at the input level rather than per camera as in FG$^2$: we place each camera in its $90^\circ$ azimuth sector $\delta_c \in \{0^\circ,90^\circ,180^\circ,270^\circ\}$ and concatenate them in order (front, right, back, left) into a single equirectangular-style panorama. The aerial branch receives the DOP20 tile resized to CCVPE's $512\times512$ input and kept north-up. Since orientation is estimated internally, no heading prior is supplied: the network recovers the full $360^\circ$ vehicle yaw jointly with position in a single forward pass. We leave the CCVPE architecture and its training recipe unchanged. For a detailed description of their approach we refer to the original paper.

We report CCVPE only in the prior-free (north-up) setting. Unlike methods that consume a heading prior by rotating the aerial tile, CCVPE keeps the tile north-up and searches over panorama rotations, so a prior can only restrict that search. In a control experiment on the original run, a perfect heading prior left localization unchanged ($31.2$ vs.\ $31.1$\,m median on matched dev scenes): an IMU-prior variant would reproduce the north-up localization numbers while trivializing heading to the prior itself. We report CCVPE from its best-position dev checkpoint.

\section{Shi et al. baseline}
The method of Shi et al.~\cite{shi2023boosting} estimates rotation by neural optimization and translation by dense spatial correlation between a ground-view BEV projection and the aerial image. It assumes a bounded pose prior and normalizes rotation to the prior range, so it applies only to the IMU setting. We use the single front camera, its native KITTI configuration, with the real Basler intrinsics rescaled to the expected $1024\times256$ input. The aerial tile is downsampled to $512\times512$ ($0.40$\,m/px) so that the correlation search covers the full tile, and the translation search range is set to $51.2$\,m accordingly. The ground-plane homography uses a camera height of $2.0$\,m, our best estimate for the roof-mounted rig. The authors pretrained checkpoints are not publicly downloadable, so we initialize the VGG-16 backbones from ImageNet and train the full model ($7.2$\,M parameters) on our data. The network and losses are unchanged.
\section{HC-Net baseline}
HC-Net~\cite{wang2023fine} regresses a correlation-aware homography between a spherically projected ground panorama and the aerial patch, giving sub-pixel position and orientation. We feed it the same four-camera pseudo-panorama as CCVPE, projected to a $512\times512$ quasi-BEV with HC-Net's own spherical transform, and the DOP20 tile resized to $512\times512$. We initialize from the released VIGOR checkpoint and fine-tune the full network ($11.3$\,M parameters). One adaptation is required for our setting: HC-Net's orientation loss assumes residuals within $\pm45^\circ$, so we wrap the orientation residual to $\pm180^\circ$ for north-up training. In the IMU setting the prior enters by pre-rolling the pseudo-panorama toward north. HC-Net has no orientation search to restrict. Everything else follows the authors recipe.

\section{Implementation details}
All learned baselines and all variants of our model are trained with three seeds and tables report mean and standard deviation over seeds. In the IMU setting, every method is evaluated with the same deterministic per-scene prior error of up to $\pm10^\circ$.
  
For FG$^2$ we fine-tune from the authors' KITTI checkpoint for $30$ epochs at effective batch $16$ using Adam with a learning rate $10^{-5}$, keeping the original InfoNCE and virtual-correspondence losses. Ground images are resized to $364\times1232$ and the aerial tile to $504\times504$. We apply no image augmentation, the four-camera variant samples one surround camera per scene during training. FG$^2$ updates $60.6$\,M parameters. For Shi et al.\ we train $15$ epochs at batch $4$ with Adam, learning rate $10^{-4}$ dropping to $10^{-5}$ after the first epoch, following the authors KITTI schedule. For HC-Net we train $30$ epochs at batch $16$ with AdamW under a one-cycle schedule with a peak learning rate $2\times10^{-4}$, weight decay $5\times10^{-4}$, gradient clipping $1.0$, six GRU refinement iterations, following the authors VIGOR recipe. CCVPE trains for $15$ epochs at batch $8$ with Adam at $10^{-4}$.
  
Our model trains for $30$ epochs at effective batch $16$ with AdamW under a cosine schedule with one warm-up epoch, using a learning rate of $10^{-3}$ for the projection layers and $5\times10^{-4}$ for the heads, weight decay of $10^{-4}$ and parameters without weight dimensions (norms, embeddings, the temperature, and the per-modality gains) are excluded from weight decay. The only augmentation is applied to the LiDAR point cloud: random dropout of up to $20\%$ of the points, per-point Gaussian jitter ($\sigma=0.05$\,m), global isotropic scaling of $\pm3\%$, and a random $\pm2^\circ$ yaw rotation. The point cloud is subsampled to $8{,}000$ points. The LoRA adapter on Utonia uses rank $8$ and $\alpha=16$, adding $1.05$\,M trainable parameters. During training we apply per-sample modality dropout in the fusion: with probability $0.3$ the cameras are dropped, with probability $0.15$ the LiDAR is dropped, and with probability $0.55$ both modalities are kept. In total the trainable surface is $3.6$\,M parameters for the image-only model, $6.2$\,M for the LiDAR-only variant, and $7.4$\,M with both branches of which the fusion gate accounts for roughly $300$ parameters, an order of magnitude below FG$^2$'s $60.6$\,M. Experiments ran on a DGX-2 with $16$ V100 $32$\,GB GPUs. Our model trains data-parallel across eight GPUs, each baseline on a single GPU.

\section{Additional ablations}
\subsection{Robustness to the heading prior}
The IMU setting rotates the aerial tile by the supplied heading prior before matching, so the model only has to resolve a bounded residual $|\Delta\theta|\le10^\circ$. \Cref{tab:robustness} shows how accurate that prior must actually be: at test time we corrupt it by a fixed offset and re-evaluate. Localization is robust for all three variants, staying within ${\sim}1$\,m of the perfect-prior value out to $\pm20^\circ$ and losing only ${\sim}3$\,m at a gross $\pm45^\circ$ offset. Orientation is, as expected, more sensitive: the classifier holds a few degrees while the prior is good to $\pm10^\circ$ and rises to ${\sim}18^\circ$ once the true residual leaves the trained range. Three points follow: first, the model needs only a coarse heading prior to realize most of the localization benefit. Second the fusion keeps its full margin over LiDAR-only at every prior quality and third all three variants degrade at the same rate, so prior tolerance does not depend on the sensor mix.

\begin{table}
      \centering
      \resizebox{\linewidth}{!}{
      \begin{tabular}{c cc cc cc}
        \toprule
        & \multicolumn{2}{c}{Ours (RGB)} & \multicolumn{2}{c}{Ours (RGB+L)} & \multicolumn{2}{c}{Ours (LiDAR-only)}
  \\
        \cmidrule(lr){2-3}\cmidrule(lr){4-5}\cmidrule(lr){6-7}
        Prior $|\Delta\theta|$ & Loc & Ori & Loc & Ori & Loc & Ori \\
        \midrule
        $0^\circ$  & 21.3 & 4.6  & \textbf{13.7} & 4.6  & 16.2 & \textbf{0.3} \\
        $5^\circ$  & 21.1 & 3.1  & \textbf{13.9} & 3.0  & 16.2 & 2.7  \\
        $10^\circ$ & 20.9 & 3.9  & \textbf{14.4} & 3.8  & 16.4 & 3.5  \\
        $20^\circ$ & 21.8 & 6.9  & \textbf{14.8} & 6.4  & 17.1 & 6.8  \\
        $45^\circ$ & 24.0 & 18.4 & \textbf{17.1} & 18.2 & 19.7 & 18.8 \\
        \bottomrule
      \end{tabular}
      }
      \caption{\textbf{Robustness to heading-prior error (test, median; Loc
      in m, Ori in $^\circ$; seed-0 checkpoints).} As the injected prior
      error $|\Delta\theta|$ grows, localization stays nearly flat up to
      $\pm20^\circ$ and loses only ${\sim}3$\,m at $\pm45^\circ$;
      orientation stays at a few degrees up to $\pm10^\circ$ and degrades
      gracefully beyond ($0^\circ$ = perfect prior). A coarse prior
      suffices, and the fusion advantage persists at every prior quality.}
      \label{tab:robustness}
\end{table}

\section{Self-similarity of aerial tiles}
\label{app:selfsim}
For each test scene we measure how many locations in the aerial tile look like the true one. We take frozen, non-finetuned DINOv3 ViT-L/16 patch tokens of the north-up $1024^2$ tile ($64\times64$ patches of $3.2\times3.2$\,m), pool a $3\times3$ descriptor at the GT position, and  compute its cosine similarity to every patch outside a $16$\,m exclusion  window. A patch with similarity above a threshold $\tau$ counts as a  look-alike. The probe shares no weights with any evaluated method except our own model's frozen encoder.
\begin{table}[h]
    \centering
    \resizebox{\linewidth}{!}{%
    \begin{tabular}{l cc}
      \toprule
      & Paved & Unpaved  \\
      \midrule
      Look-alike positions ($\tau{=}0.8$)     & $79$ ($1.9\%$) & $101$ ($2.5\%$)  \\
      Look-alike fraction ($\tau{=}0.7$)      & $5.6\%$ & $7.8\%$ \\
      \bottomrule
    \end{tabular}%
    }
    \caption{\textbf{Perceptual aliasing by terrain} (test, per-scene medians, unpaved $>$ paved). Unpaved scenes contain $1.28\times$ more look-alike positions at $\tau{=}0.8$.}
    \label{tab:selfsim_terrain}
\end{table}
  
\begin{table}[h]
    \centering
    \resizebox{0.85\linewidth}{!}{%
    \begin{tabular}{l cccc}
      \toprule
      & Winter & Spring & Summer & Autumn \\
      \midrule
      Look-alike positions ($\tau{=}0.8$) & $63$ & $79$ & $117$ & $102$ \\
      \bottomrule
    \end{tabular}%
    }
    \caption{\textbf{Perceptual aliasing by season} (median per-scene look-alike count). Aliasing peaks in the leaf-on seasons, matching where the appearance-based baselines degrade most.}
    \label{tab:selfsim_season}
\end{table}

\begin{table}[h]
    \centering
    \resizebox{\linewidth}{!}{%
    \begin{tabular}{ll ccc}
      \toprule
      Method & Setting & $\rho$ (all) & $\rho$ (paved) & $\rho$ (unpaved) \\
      \midrule
      FG$^2$ (surround) & IMU & $+0.23$ & $+0.18$ & $+0.20$ \\
      Shi et al.\ (front) & IMU & $+0.21$ & $+0.19$ & $+0.20$ \\
      Ours (RGB+L)      & IMU & $+0.18$ & $+0.26$ & $+0.05$ \\
      CCVPE (surround)  & NA  & $+0.16$ & $+0.17$ & $+0.10$ \\
      Ours (LiDAR-only) & IMU & $+0.15$ & $+0.23$ & $+0.01$ \\
      FG$^2$ (surround) & NA  & $+0.13$ & $+0.08$ & $+0.10$ \\
      Ours (RGB)        & IMU & $+0.10$ & $+0.15$ & $+0.02$ \\
      Ours (RGB+L)      & NA  & $+0.08$ & $+0.12$ & $-0.00$ \\
      HC-Net (surround) & IMU & $+0.07$ & $+0.05$ & $+0.05$ \\
      Ours (LiDAR-only) & NA  & $+0.05$ & $+0.10$ & $-0.05$ \\
      Ours (RGB)        & NA  & $+0.05$ & $+0.12$ & $-0.04$ \\
      HC-Net (surround) & NA  & $+0.04$ & $+0.01$ & $+0.03$ \\
      \bottomrule
    \end{tabular}%
    }
    \caption{\textbf{Rank correlation between per-scene aliasing and localization error} (Spearman $\rho$, look-alike count at $\tau{=}0.8$, averaged over training seeds). Every method correlates positively. The matching-based baselines correlate most and remain positive within both terrain groups. Our fusion is least affected in the prior-free setting.}
    \label{tab:selfsim_corr}
\end{table}

Localization error rises with aliasing for every method. The correlation is strongest for the matching-based baselines FG$^2$ and Shi et al.. Correlations are weaker in the north-up setting, where orientation ambiguity dominates the position error, and weakest for HC-Net, whose errors stem mainly from its ground-plane homography assumption. The seasonal profile mirrors the localization results: aliasing peaks in summer, exactly where the LiDAR advantage over RGB is largest.
\section{Aerial imagery statistics}
\label{app:aerial}
All aerial tiles are official DOP20 digital orthophotos of the Bavarian survey administration (Bayerische Vermessungsverwaltung), published as open data under CC\,BY~4.0. Each scene uses a north-aligned $1024\times1024$ crop at $0.20$\,m ground sampling distance, covering $204.8\times204.8$\,m centered near the scene position. The underlying survey flights all took place in the leaf-on season (June to August), so the overhead view is summer-only regardless of the season in which the ground data was recorded; Tab.~\ref{tab:aerial_dates} lists the capture dates per split. For a small number of train and dev scenes ($33$ and $26$) the capture date could not be resolved from the provider metadata.

\begin{table}[h]
    \centering
    \begin{tabular}{l ccc}
      \toprule
      Capture date & train & dev & test \\
      \midrule
      2024-06-18 & $550$  & $72$  & $3$    \\
      2024-07-31 & $1590$ & $177$ & $326$  \\
      2024-08-24 & $3920$ & $749$ & $1517$ \\
      2025-08-19 & $27$   & $3$   & $16$   \\
      2025-08-26 & $310$  & $37$  & $0$    \\
      \bottomrule
    \end{tabular}%
    
    \caption{\textbf{DOP20 orthophoto capture dates per split.} All survey flights fall in the leaf-on season. The two most recent dates stem from the 2025 flight campaign of the same areas.}
    \label{tab:aerial_dates}
\end{table}

\section{Buffered test split}
\label{app:buffered}
Scenes were sampled along trajectories, so some test tiles share overhead imagery with training tiles: $34\%$ of test scenes lie within one tile side ($204.8$\,m) of a training scene. The buffered test split removes this overlap. A test scene is retained only if its aerial tile provably shares no pixel with the tile of any training scene, using a conservative per-axis bound on the tile-center distance that accounts for the per-scene offset between vehicle position and tile center. Thus, borderline pairs count as overlapping. This leaves $840$ of $1862$ scenes with a median distance of $481$\,m to the nearest training scene (full test: $252$\,m). 

\begin{table}[h]
    \centering
    \resizebox{\linewidth}{!}{%
    \begin{tabular}{ll cc cc}
      \toprule
      & & \multicolumn{2}{c}{Full test ($n{=}1862$)} & \multicolumn{2}{c}{Buffered ($n{=}840$)} \\
      \cmidrule(lr){3-4} \cmidrule(lr){5-6}
      Setting & Method & Loc\,(m) & Ori\,($^\circ$) & Loc\,(m) & Ori\,($^\circ$) \\
      \midrule
      North-up & CCVPE (surround)  & 45.4{\scriptsize$\pm$2.9} & 35.5{\scriptsize$\pm$17.0} & 46.3{\scriptsize$\pm$2.8} & 39.9{\scriptsize$\pm$13.0} \\
               & HC-Net (surround) & 31.8{\scriptsize$\pm$0.4} & 84.4{\scriptsize$\pm$1.2}  & 31.8{\scriptsize$\pm$0.2} & 83.6{\scriptsize$\pm$2.1} \\
               & FG$^2$ (surround) & 34.4{\scriptsize$\pm$0.4} & 93.1{\scriptsize$\pm$0.3}  & 34.2{\scriptsize$\pm$0.2} & 92.8{\scriptsize$\pm$0.2} \\
               & Ours (RGB)        & 29.8{\scriptsize$\pm$0.1} & 27.9{\scriptsize$\pm$3.3}  & 28.8{\scriptsize$\pm$0.1} & 35.1{\scriptsize$\pm$10.6} \\
               & Ours (LiDAR-only) & 28.7{\scriptsize$\pm$0.3} & 38.5{\scriptsize$\pm$5.4}  & 29.2{\scriptsize$\pm$0.4} & 49.3{\scriptsize$\pm$6.4} \\
               & Ours (RGB+L)      & \textbf{27.1{\scriptsize$\pm$0.3}} & \textbf{24.8{\scriptsize$\pm$0.3}} & \textbf{26.2{\scriptsize$\pm$0.7}} & \textbf{29.6{\scriptsize$\pm$1.8}} \\
      \midrule
      IMU $\pm10^\circ$ & Shi et al.\ (front) & 35.5{\scriptsize$\pm$0.5} & \textbf{3.1{\scriptsize$\pm$0.2}} & 34.1{\scriptsize$\pm$0.4} & \textbf{3.5{\scriptsize$\pm$0.1}} \\
               & HC-Net (surround) & 31.9{\scriptsize$\pm$3.3} & 11.7{\scriptsize$\pm$0.8} & 33.8{\scriptsize$\pm$6.6} & 12.5{\scriptsize$\pm$0.6} \\
               & FG$^2$ (surround) & 25.0{\scriptsize$\pm$0.4} & 4.4{\scriptsize$\pm$0.2}  & 25.2{\scriptsize$\pm$0.5} & 4.9{\scriptsize$\pm$0.3} \\
               & Ours (RGB)        & 25.1{\scriptsize$\pm$0.1} & 3.9{\scriptsize$\pm$0.1}  & 24.5{\scriptsize$\pm$0.2} & 4.2{\scriptsize$\pm$0.2} \\
               & Ours (LiDAR-only) & 21.2{\scriptsize$\pm$0.2} & 3.5{\scriptsize$\pm$0.1}  & 20.9{\scriptsize$\pm$0.1} & 3.8{\scriptsize$\pm$0.1} \\
               & Ours (RGB+L)      & \textbf{19.5{\scriptsize$\pm$0.3}} & 3.8{\scriptsize$\pm$0.1} & \textbf{18.7{\scriptsize$\pm$0.2}} & 4.0{\scriptsize$\pm$0.2} \\
      \bottomrule
    \end{tabular}%
    }
    \caption{\textbf{Full vs.\ buffered test} (Loc = mean position error, Ori = median orientation error, $\pm$ std over 3 seeds). No method gains from the removed overlap: every ranking persists on the leakage-free subset, the paired fusion advantage grows (cf.\ the main text), and orientation degrades most for the camera-free variant.}
    \label{tab:buffered_full}
\end{table}

\begin{table}[h]
    \centering
    \resizebox{0.9\linewidth}{!}{%
    \begin{tabular}{l ccc c}
      \toprule
      Pair & Min & 5th pct.\ & Median & Tile overlap \\
      \midrule
      test $\to$ train & $1.5$\,m & $30$\,m  & $252$\,m & $40\%$ \\
      dev $\to$ train  & $0.4$\,m & $20$\,m  & $179$\,m & $58\%$ \\
      test $\to$ dev   & $0.1$\,m & $10$\,m  & $141$\,m & $66\%$ \\
      \bottomrule
    \end{tabular}%
    }
    \caption{\textbf{Inter-split nearest-neighbor distances} between scene positions, and the share of scenes whose $204.8$\,m aerial crop overlaps a crop of the other split.}
    \label{tab:intersplit}
\end{table}

\end{document}